\documentclass[conference]{IEEEtran}
\IEEEoverridecommandlockouts
\usepackage{cite}
\usepackage{amsmath,amssymb,amsfonts}
\usepackage{algorithmic}
\usepackage{graphicx}
\usepackage{textcomp}
\usepackage{xcolor}
\usepackage{blindtext}
\usepackage{url}
\usepackage{comment}
\usepackage[skip=0pt]{caption}

\def\BibTeX{{\rm B\kern-.05em{\sc i\kern-.025em b}\kern-.08em
    T\kern-.1667em\lower.7ex\hbox{E}\kern-.125emX}}
\begin{document}
%\addbibresource{references.bib}

\title{Towards A Reliable Ground-Truth For Biased Language Detection
\thanks{This work was supported by the German Academic Exchange Service and the Hanns-Seidel-Foundation.}
}

%\author{Anonymous authors}
\author{
\IEEEauthorblockN{Timo Spinde}
\IEEEauthorblockA{
\small
\textit{Dept. of Computer and} \\
\textit{Information Science} \\
\textit{University of Konstanz}\\
Constance, Germany \\
timo.spinde@uni-konstanz.de}

\and
\IEEEauthorblockN{David Krieger}
\IEEEauthorblockA{
\small
\textit{Dept. of Computer and} \\
\textit{Information Science} \\
\textit{University of Konstanz}\\
Constance, Germany \\
jan-david.krieger@uni-konstanz.de}
\and
\IEEEauthorblockN{Manuel Plank}
\IEEEauthorblockA{
\small
\textit{Dept. of Computer and} \\
\textit{Information Science} \\
\textit{University of Konstanz}\\
Constance, Germany \\
manuel.plank@uni-konstanz.de}
\and
\IEEEauthorblockN{Bela Gipp}
\IEEEauthorblockA{
\small
\textit{School of Electrical, Information} \\
\textit{ and Media Engineering} \\
\textit{University of Wuppertal}\\
Wuppertal, Germany \\
gipp@uni-wuppertal.de}
}

\maketitle

\begin{abstract}
Reference texts such as encyclopedias and news articles can manifest biased language when objective reporting is substituted by subjective writing. Existing methods to detect bias mostly rely on annotated data to train machine learning models. However, low annotator agreement and comparability is a substantial drawback in available media bias corpora. 
%Even our specifically designed survey platforms do not yield sufficient annotator accordance. 
%Den Satz hab ich jetzt rausgenommen, weil wir sonst TASSY noch im Text beschreiben müssten
To evaluate data collection options, we collect and compare labels obtained from two popular crowdsourcing platforms. Our results demonstrate the existing crowdsourcing approaches' lack of data quality, underlining the need for a trained expert framework to gather a more reliable dataset. By creating such a framework and gathering a first dataset, we are able to improve Krippendorff's $\alpha$ = 0.144 (crowdsourcing labels) to $\alpha$ = 0.419 (expert labels). We conclude that detailed annotator training increases data quality, improving the performance of existing bias detection systems. We will continue to extend our dataset in the future.
%hab jetzt 0.101 rausgelassen, weil man ja vom besseren ausgehen kann für den vergleich zu experts
\end{abstract}

\begin{IEEEkeywords}
Media Bias, News Slant, Dataset, Survey, Crowdsourcing 
\end{IEEEkeywords}

\section{Introduction and Related Work}
The way journalists report on newsworthy events can influence consumers in their perception of political issues. Slanted coverage, also known as \textit{media bias}, appears in different forms and on various linguistic levels \cite{Spinde2021c}. The present project deals with the exploration of biased language on a word and sentence level.

Several studies have presented systems to detect slanted news reporting. Efforts include traditional machine learning classifiers relying on manual feature-engineering as in \cite{spinde2021automated,Spinde2020}, and neural-based methods \cite{hube2019neural}. To train and evaluate these algorithms, instances of text with a bias-inducing word choice or framing need to be labeled \cite{Spinde2021b}. The need for training and validation data can be addressed by designing a crowdsourcing task as in \cite{Spinde2021MBIC}. The dataset created in this study via \textit{Amazon Mechanical Turk} (MTurk) is the most exhaustive sample containing news bias labels on a fine-grained level to the best of our knowledge.\footnote{\url{https://www.mturk.com/}} Yet, one of the approaches' main shortcomings is the resulting poor data quality concerning inter-rater reliability (IRR), which might negatively affect the performance of downstream classification tasks. Machine learning algorithms need rich training signals to learn an accurate language representation.

Crowdsourcing via MTurk has shown several drawbacks in past research: known problems are practice effects and the existence of discussion boards, resulting in a reduced naivety of the users. In contrast, those problems are not found to be existent to the same extent on other crowdsourcing platforms such as  \textit{Prolific} \cite{palan2018prolific}.\footnote{\url{https://www.prolific.com/}} In further comparative annotation studies, MTurkers were less naïve and
more familiar with the presented tasks than Prolific users. Beyond that, MTurkers showed a higher cheating rate than Prolific participants \cite{peer2017beyond}. 

Our work aims to facilitate further research on the language conveying bias by elaborating on different ways to get fine-grained and qualitative annotations of biased language. As a first step towards compiling a reliable ground-truth for biased language detection, we compare IRR scores regarding media bias annotations on both MTurk and Prolific. Thereupon, we let trained experts label bias instances using detailed annotation instructions. 

We summarize our hypotheses as follows:
We assume that Prolific crowdsourcers show a higher agreement than MTurkers due to the presented drawbacks of the MTurk platform. Beyond that, we presume that expert training through detailed labeling instructions increases the annotation accordance.

\section{Methodology and Results}

\subsection{Crowdsourcing}
We first seek to compare user performances on MTurk and Prolific in the context of media bias. We rely on the news bias dataset provided by \cite{Spinde2021MBIC} as data ground-truth. It comprises 1,700 sentences with news bias annotations on word and sentence level extracted from 1000 articles. The dataset covers news platforms from the whole political spectrum. Furthermore, the survey includes a wide range of controversial topics with a balanced sociodemographic user characteristics distribution. The dataset being representative is crucial for the development of a generalizable bias detection tool.

\begin{figure*}[t!]
    \centering
    \def\svgwidth{400pt} % \columnwidth
    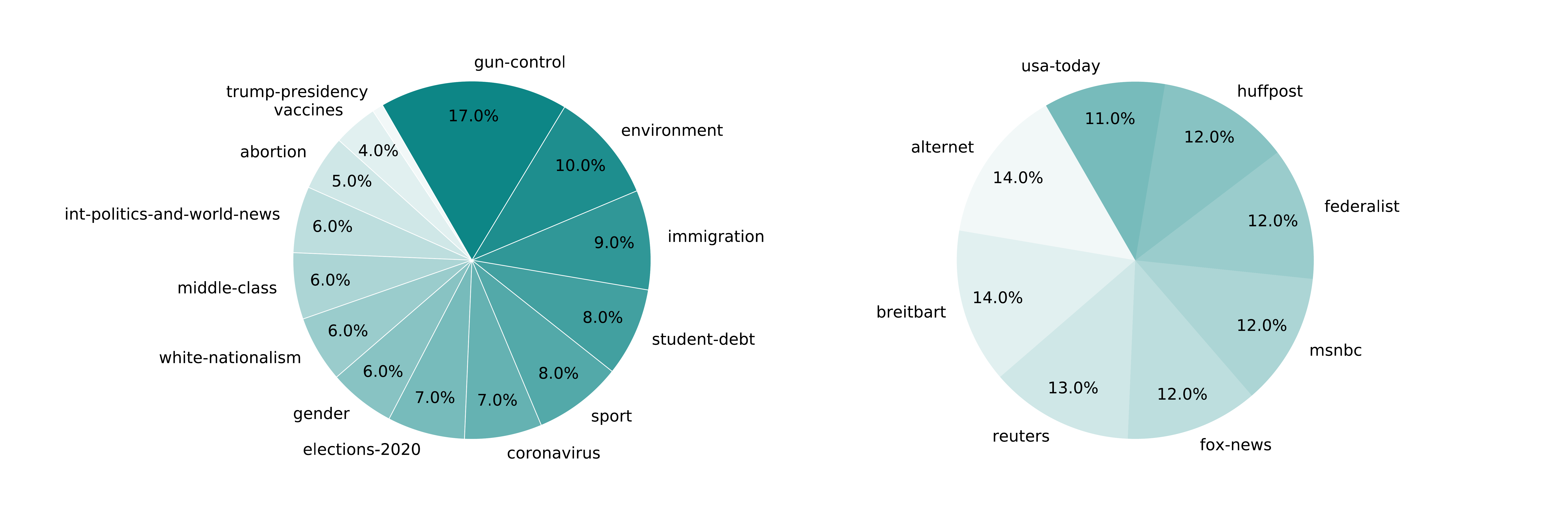
    \caption{Topic (left) and News Outlet (right) Distribution}
\end{figure*}

We draw a representative sample of 100 sentences from the existing dataset \cite{Spinde2021MBIC}. The sample's characteristics are illustrated in Fig. 1. To obtain a clear comparative picture we compare performance on MTurk and Prolific by asking Prolific users to re-annotate the existing MTurk labels. As in the original study, we ask the annotators to label the sentences in terms of bias on a sentence level and a word level. Our agreement metric of choice is Krippendorff's $\alpha$ \cite{krippendorff2004content}. The computational principle of this metric involves the ratio of the observed disagreement against the expected disagreement. We focus on the agreement on sentence level since including one linguistic level suffices for our crowdsourcing comparison. In future approaches, we also aim to analyze user agreement on a more fine-grained level.
The agreement in the original study conducted on MTurk was $\alpha = .101$. Annotators on Prolific reach an $\alpha = .144$. The average number of clickworkers were 10.43 and 12.69 in the MTurk and Prolific study, respectively.

\begin{comment}
\begin{table}[htbp]
\caption{Results of crowdsourcing comparison}
\begin{center}
\begin{tabular}{|p{3,5cm}|c|c|}
\hline
\textbf{Metrics} & \textbf{MTurk Sample}& \textbf{Prolific Sample}\\
\hline
Krippendorff's $\alpha$ & 0.101 & 0.144 \\
\hline
Number of sentences & 100 & 100 \\
\hline
\textcolor{red}{Annotators per sentence $\varnothing$} & 10.43 & 12.69 \\
\hline
\end{tabular}
\label{tab1}
\end{center}
\end{table}
\end{comment}

The Profilic users show a more profound agreement overlap in their bias ratings than the MTurkers. Yet, both agreement scores are by far not satisfying since \cite{krippendorff2004content} suggests a minimum  $\alpha$ = .667 as the lowest conceivable limit. No available dataset comes near to that margin. These findings support the notion that identifying biased language is a complex task. We assume that crowdsourcers do not have sufficient knowledge regarding the linguistic theory and manifestations of the media bias concept. As a logical next step, we want to include bias ratings of experts to enhance data quality.

\subsection{Expert Annotations}
We hypothesize that the low quality of the data is the result of limited time. Crowdsourcers might mostly not be able to render accurate labels for this complex task. To mitigate scant domain knowledge problems, we follow corpus linguistic practice and develop detailed annotation instructions for coders. Annotation guidelines and data can be found at \url{https://zenodo.org/record/4625151}. 
Providing methodological steps for human-coders is essential but cannot be addressed sufficiently in a crowdsourcing setting due to time constraints. For the first sample presented in this poster, we employ two annotators working in the context of media bias. Preliminary results on 1,700 sentences are encouraging, yielding an $\alpha$ of .419, which surpasses available datasets by a large margin. Exceeding the annotator pool to 12 annotators and experimenting with other users is a work-in-progress.

\section{Conclusions and Future Work}
This poster proposes a work-in-progress approach to compile a ground-truth dataset suitable for media bias detection.
So far, we implemented a crowdsourcing comparison of user performances on the media bias detection task. We let users annotate an exemplary sentence corpus both on MTurk and Prolific. Prolific participants outperformed MTurkers with a Krippendorff's $\alpha$ = .144 vs. $\alpha$ = .101. We conclude that these low agreement scores are due to the crowdsourcers' insufficient understanding of the media bias concept. 

Furthermore, including trained experts improved the data quality by increasing the annotators' agreement to an $\alpha$ = .419. As a next step, we plan to build a diverse team of annotators to improve the current dataset quantitatively and qualitatively. We expect that future computational detection approaches will benefit substantially from this development.

\bibliographystyle{unsrt}
\bibliography{short.bib}

\begin{comment}

\end{comment}

\end{document}